\documentclass[10pt,twocolumn,letterpaper]{article}

\usepackage{wacv}              % To produce the CAMERA-READY version
\usepackage{graphicx}
\usepackage{amsmath}
\usepackage{amssymb}
\usepackage{booktabs}
\usepackage{cite}
\usepackage{bm}
\usepackage{xcolor}
\usepackage{algorithm}
\usepackage{algorithmic}
\usepackage{subcaption}
\usepackage{textcomp} % Provides \textdegree
\DeclareUnicodeCharacter{00B0}{\textdegree} % Degree symbol declaration
\usepackage[pagebackref,breaklinks,colorlinks]{hyperref}

\usepackage[capitalize]{cleveref}
\crefname{section}{Sec.}{Secs.}
\Crefname{section}{Section}{Sections}
\Crefname{table}{Table}{Tables}
\crefname{table}{Tab.}{Tabs.}

\def\wacvPaperID{1486} % *** Enter the WACV Paper ID here
\def\confName{WACV}
\def\confYear{2025}

\DeclareMathOperator*{\argmax}{arg\,max}

\begin{document}

\author{
    Roman Colman\textsuperscript{1}, Minh Vu\textsuperscript{2}, Manish Bhattarai\textsuperscript{2}, Martin Ma\textsuperscript{1},\\
    Hari Viswanathan\textsuperscript{1}, Daniel O'Malley\textsuperscript{1}, Javier E. Santos\textsuperscript{1} \\[0.5em]
    \textsuperscript{1}\textit{Earth and Environmental Sciences, Los Alamos National Laboratory, Los Alamos, NM, USA} \\[0.3em]
    \textsuperscript{2}\textit{Theoretical Division, Los Alamos National Laboratory, Los Alamos, NM, USA}
}

\title{PatchFinder: Leveraging Visual Language Models for Accurate Information Retrieval using Model Uncertainty}

\maketitle

\begin{abstract}

For decades, corporations and governments have relied on scanned documents to record vast amounts of information. However, extracting this information is a slow and tedious process due to the sheer volume and complexity of these records. The rise of Vision Language Models (VLMs) presents a way to efficiently and accurately extract the information out of these documents. The current automated workflow often requires a two-step approach involving the extraction of information using optical character recognition software, and subsequent usage of large language models for processing this information. Unfortunately, these methods encounter significant challenges when dealing with noisy scanned documents, often requiring computationally expensive language models to handle high information density effectively.

In this study, we propose \textit{PatchFinder}, an algorithm that builds upon VLMs to improve information extraction. First, we devise a confidence-based score, called \textit{Patch Confidence}, based on the Maximum Softmax Probability of the VLMs' output to measure the model's confidence in its predictions. Using this metric, PatchFinder determines a suitable patch size, partitions the input document into overlapping patches, and generates confidence-based predictions for the target information.
Our experimental results show that PatchFinder, leveraging Phi-3v, a 4.2 billion parameter VLM, achieves an accuracy of 94\% on our dataset of 190 noisy scanned documents, outperforming ChatGPT-4o by 18.5 percentage points.
\end{abstract}

\section{Introduction}

For many decades, scanned documents have been a common practice for data collection by governments and other agencies. These documents contain valuable information spanning a wide range of topics \cite{nara_digitization_2015}. For example, information in these documents includes details about weather patterns, financial records of companies, details about historical civilizations, law records, medical records, and much more. Traditionally, extracting this information involved humans identifying and transferring the data into databases, resulting in substantial costs and potential human error. Naturally, developing an automated method for information extraction has become a high priority task among researchers in computer vision \cite{dhouib2023docparserendtoendocrfreeinformation, yu2020pickprocessingkeyinformation, kim2022ocrfreedocumentunderstandingtransformer}, Optical Character Recognition (OCR) \cite{perot2024lmdxlanguagemodelbaseddocument, Ma2024}, and data management \cite{10.1007/978-981-19-2600-6_4}. The most common method of automatically extracting information from scanned documents is OCR, which utilizes pattern matching and feature extraction to convert images of text into machine-readable text formats \cite{4376991}. The recent advance of large language models (LLMs) additionally allows novel tools to further process the OCR-extracted information into more structured answers about a target piece of information \cite{Ma2024}. However, high-quality OCR tools capable of accurately extracting information from millions of scanned documents are often very expensive, posing a significant challenge for individuals and organizations with limited funding. The two-step approach can also introduce error propagation at each step, making it inflexible to various layouts and noise levels and also is potentially constrained by a model's context length. \cite{Ma2024, perot2024lmdxlanguagemodelbaseddocument}. Additionally, as existing methods are not specifically designed for this task,  their performance is sub-optimal (See Table~\ref{tab:benchmarks}).

Vision Language Models (VLMs) present an opportunity to remedy many of the issues related to traditional OCR methods due to their strong presence in the open-source community, as well as their end-to-end workflow. These VLMs use both a vision and a language component internally, allowing for a more efficient approach. In particular, this work contributes a method named \textit{PatchFinder} to utilize VLMs to address the task of extracting information from scanned documents. PatchFinder operates based on our newly proposed \textit{Patch Confidence} score, which measures the VLM's confidence in its predictions. Then, PatchFinder uses the PC score to examine the training data and determine an appropriate patch size. After that, PatchFinder 
divides the input image into patches of that size, evaluates the confidence of each patch's prediction, and returns the final prediction based on the Patch Confidence score. 

By integrating both visual and 
text information effectively, our approach demonstrates significant performance improvements over existing methods. This work proves the potential of VLMs to revolutionize document analysis tasks, particularly for complex and noisy documents.

\begin{table}[htbp]
\resizebox{\linewidth}{!} 
{
    \centering
    \begin{tabular}{lcccc}
        \toprule
        \textbf{Model} & \textbf{Colorado} & \textbf{Pennsylvania} & \textbf{No. Params.} \\ 
        \midrule
        {LLaVa-Next} & 0.0\% & 0.0\% & 7B \\ 
        {Idefics2} & 36.7\% & 6.7\% & 8B \\ 
        {Donut} & 83.3\% & 26.7\% & 144M \\ 
        {Phi-3v} & 100.0\% & 33.3\% & 4.2B \\ 
        {OCR + Phi-3-mini} & 100.0\% & 66.7\% & \ 3.8B \\ 
        \midrule
        \textbf{PatchFinder} & \textbf{100.0\%} & \textbf{93.3\%} & 4.2B \\ 
        \bottomrule
    \end{tabular}
    }
    \caption{Preliminary results of different methods on 10 documents from Colorado and Pennsylvania datasets.}
    \label{tab:benchmarks}
\end{table}
\section{Related Work and Benchmarks} \label{sect:prelim}

\subsection{Related Work}

Information extraction from scanned documents, especially those with complex layouts, has been a grand challenge. Traditional OCR methods, while useful for clean and well-structured texts, often struggle with noisy backgrounds, varied fonts, and handwritten content. These traditional methods also require a processing step, done by LLMs, in order to format the information into a usable structure. This two-step requirement leads to context loss and inefficiencies, indicating that an end-to-end approach would be a more effective way to extract this information. In this study, we propose VLMs as a critical solution to these issues with the traditional methods of information extraction.

Recent VLMs leverage large-scale image-text pairs for pre-training, enabling them to learn rich vision-language representations. For instance, models like CLIP~\cite{radford2021learning} and BLIP-2~\cite{li2023blip} utilize vision transformers (ViTs)~\cite{vit} combined with language models to interpret and process visual data in a context-aware manner. These models have shown remarkably effective in tasks such as image captioning, visual question answering, and scene text recognition~\cite{zhang2024vision,rotstein2024fusecap}. By integrating visual features with contextual language models like BERT~\cite{devlin2018bert} and GPT~\cite{brown2020language}, VLMs can handle diverse document layouts and degraded text quality quite effectively.~\cite{zhu2023minigpt}.

The ability to find and interpret the confidence in the response of a model would be a massive tool for interpreting responses and preventing hallucinations. One of the most researched proxies for this confidence of a model is called the Maximum Softmax Probability (MSP). The MSP calculates the highest probability from the model’s output distribution of each output token, and has been validated in various studies, including those by Hendrycks and Gimpel \cite{hendrycks2018baselinedetectingmisclassifiedoutofdistribution}, who demonstrated that MSP could serve as a possible method of flagging errors.
Enhancements such as thresholding techniques to flag low-confidence outputs and combining MSP with other uncertainty measures have been explored to improve reliability~\cite{taha2022confidenceestimationclassificationbased}.

Our study builds on these advancements by introducing the PatchFinder algorithm, which utilizes Patch Confidence (PC) to improve information extraction processes. By dividing the input document into smaller patches and evaluating the confidence of each patch's predictions using MSP, PatchFinder focuses on the most reliable parts of the image. 

\subsection{Scope of Work and Setup} 

A key contribution in the broad scope of this work is in introducing our PatchFinder algorithm in order to aid the efforts of finding and plugging undocumented orphan wells \cite{catalog}. Such wells are known to leak methane into the environment, posing a strong environmental risk, but the number and location of many of these wells remains unknown. It is believed that the documentation of a large number of these wells lies in some of the millions of scanned well documents in existence. For the purposes of this study, we were able to access 190 scanned well documents to develop and test our PatchFinder method. See Sect. \ref{subsect:exp_setting} for more details.

More concretely, given a historical well record, our objective is to accurately extract three key pieces of information: latitude, longitude, and True Vertical Depth (TVD). 

Latitude and longitude are essential for determining the exact location of possible abandoned oil and gas wells, which helps in identifying and monitoring them. True Vertical Depth (TVD) refers to the vertical distance from the surface down to the deepest point of the well, providing a clear measure of how deep the well is drilled underground, this is important for assessing potential environmental impacts (to nearby aquifers) and planning any necessary remediation efforts (plugging operations).

The precision of these extracted values is crucial, as even small inaccuracies in latitude and longitude can lead to significant location discrepancies between wells. 

 To evaluate the information extraction process, we calculate the average accuracy in extracting the target fields, i.e., latitude, longitude, and TVD. This ensures that the evaluation is comprehensive across all required fields and documents. 

\subsection{Preliminary Benchmarks}
 
 In order to get a baseline on how open-source VLMs perform on this specific task, we leveraged LLaVa-NeXT (7B) \cite{liu2024llavanext}, Phi-3
(4.2B) \cite{abdin2024phi3technicalreporthighly}, Idefics2 (8B) \cite{Idefics2024}, and Donut (144M) \cite{kim2022ocrfreedocumentunderstandingtransformer}.
LLaVa-NeXT (7B) is a VLM known for its robust performance in multimodal tasks, combining visual and textual data to improve information extraction accuracy. The improvements over LLaVA-1.5 \cite{liu2024improvedbaselinesvisualinstruction} enabled LLaVA-NeXT to improve its accuracy on OCR-related tasks while keeping a minimalist design. 
Phi-3v (4.2B) is another VLM designed to handle document analysis tasks efficiently. It is recognized for its high accuracy in text extraction from noisy documents due to the rigorous OCR-related training that was deployed on the model, making it suitable for extracting precise information from various document types.
Idefics2 (8B) is a model that integrates advanced visual and language processing capabilities. It has demonstrated strong performance in extracting detailed and accurate information from complex documents, aligning well with our requirement for accurate data extraction.
Donut (144M) is one of the original state-of-the-art OCR-free document extraction models. It is known for its end-to-end design that is specifically focused on document extraction. Donut uses the architecture of a transformer visual encoder and a textual decoder. The model then converts target-type information into a structured JSON format for easy extraction.

As depicted in Table~\ref{tab:benchmarks}, Phi-3v exhibits noticeably better performance compared to the other models, although it achieves only 33\% accuracy on Pennsylvania documents. These results fall short of our task requirements, showcasing that there is still room for improvement when it comes to out-of-the-box models.

Despite our efforts to enhance the performance of LLaVa-NeXT and Idefics2 models, they performed poorly on this task. We experimented with various input resolutions and preprocessing techniques, prompts, and documents, but these models struggled due to the high variability in document layouts, scan qualities, and noise levels. These challenges are particularly pronounced in old documents where fonts, formats, and damages (e.g., fading, smudging) add layers of complexity that standard models are not equipped to handle effectively. The critical fields (TVD, latitude, and longitude) are often located in different parts of the documents and vary in length, which further complicates extraction tasks for traditional OCR and baseline models (refer to Fig.~\ref{fig:drilling_reports}).

We hypothesize that the accuracies between the models all come down to the training data provided to each model. It is likely that Phi-3v has been trained with training data that is more relevant to our task, and as a result, we chose it as our baseline for the remainder of the study.
Due to the popularity of the OCR two-step approach~\cite{Ma2024}, we also examine the capabilities of the Phi-3-mini language model \cite{abdin2024phi3technicalreporthighly} coupled with Google's OCR \cite{GoogleOCR}. Although OCR + Phi-3-mini outscored all other VLMs, its performance on the Pennsylvania dataset still leaves plenty of room for improvement.

\begin{figure*}[ht!]
    \centering
    \includegraphics[width=0.99\linewidth]{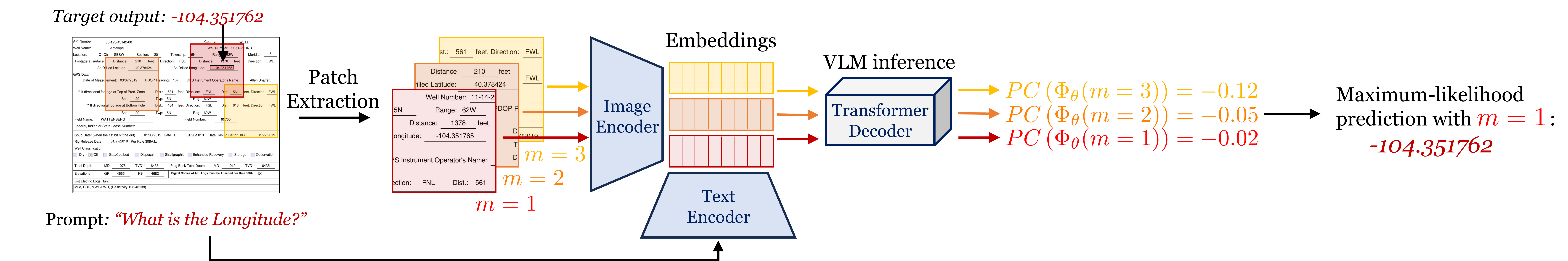}
    \caption{Illustration of PatchFinder's Confidence-Based prediction. The input image is first partitioned into multiple patches ($m=1,2,3$ are shown as examples). Then, visual tokens of each patch are obtained via an image encoder, i.e., CLIP ViT-L/14 \cite{radford2021learning}. Those tokens are then combined with the prompting-text-tokens in an interleaved way, and fed through a transformer decoder, i.e., Phi-3-mini-128K-instruct \cite{abdin2024phi3technicalreporthighly}. Using the transformer's MSP outputs, PatchFinder computes the Patch Confidence and selects the most confident patch. The target sequence is then generated with maximum-likelihood criteria on that patch. More details in Subsect.~\ref{subsect:details}.}
    \label{fig:patchfinder}
\end{figure*}
\section{Methodology} \label{sect:method}
predictions on each patch and select the most confident prediction as the final output. 
To tackle the challenges in extracting information from the historical well data (See Sect.~\ref{sect:prelim}), we devise a confidence measure, called \textit{Patch Confidence (PC)} (Subsect.~\ref{subsect:patch_confidence}), and a corresponding two part information extraction algorithm, called PatchFinder (Subsect.~\ref{subsect:patch_selection}) based on the PC to leverage VLMs for our task.  The first part of PatchFinder, called Patch Size Optimization, aims to determine the optimal patch size for the model's operations. The second part, named Confidence-Based Prediction, evaluates the model's predictions on each patch and select the most confident prediction as the final output. This structured approach, demonstrated in Fig.~\ref{fig:patchfinder}, ensures that the model can effectively process the document images by focusing on localized patches rather than the entire image at once, thereby improving its performance on historical well data. We end this section with Subsect.~\ref{subsect:details} dicussing several implementation details of PatchFinder.

\subsection{Patch Confidence} \label{subsect:patch_confidence}
We now formulate how to utilize VLMs to measure the confidence of each patch, called the Patch Confidence (PC). We begin the discussion by considering  one general objective of VLMs, which is to estimate the distribution of a sequence $\bm x$ given a context $\bm z$:
\begin{align}
    p({\bm x} | {\bm z}) \approx p_\theta({\bm x} | {\bm z}) = \prod_{i = 1}^l p_\theta(x_i|{\bm x}_{1\cdots i-1}, {\bf z})
\end{align}
Here, $\theta$ is the VLM's parameters. In our extraction problem, the context $\bm z$ includes the embedding of the prompt from the language model and the input image/patch. 

For a given window size $s$, the input image is divided into $N(s)$ patches and fed to the VLM. We denote the resulting context for each patch $m$ by $\bm z^{(m)}$, and the corresponding estimation by the VLM by $p_\theta^{(m)} := p_\theta\left({\bm x} | {\bm z^{(m)}}\right) $. Intuitively, the PC can be considered as a score computed on $p_\theta^{(m)}$ informing us on the confidence of the estimation on the patch $m$. Particularly, by expressing the estimated probability of the token $i$-th in the vector form: $p_\theta \left(x_i|{\bm x}_{1\cdots i-1}, {\bm z}^{(m)} \right) = \left[p_\theta \left(x_i^1|{\bm x}_{1\cdots i-1}, {\bm z}^{(m)} \right), \cdots, p_\theta \left(x_i^K|{\bm x}_{1\cdots i-1}, {\bm z}^{(m)} \right)]\right]$, where $K$ is the number of tokens in the vocabulary, we have the most maximum likelihood estimation for the $i$-th token is:
\begin{align}
    \hat{x}_i = x^{k_i^*}, \ \textup{where } k_i^* = \argmax_{k \in \{1,\cdots, K\} }  p_\theta \left(x_i^k|{\bm x}_{1\cdots i-1}, {\bm z}^{(m)} \right) \nonumber
\end{align}
The confidence of a patch $m$, i.e., Patch Confidence, is then defined as the average log-likelihood of the maximum-likelihood sequence:
\begin{align}
    &PC(m) = PC \left( p_\theta \left({\bm x}| {\bm z}^{(m)} \right) \right) \nonumber
    \\
    :=&\frac{1}{l}\sum_{i=1}^l \log \left( p_\theta \left(x^{k_i^*}|{\bm x}_{1\cdots i-1}, {\bm z}^{(m)} \right) \right) \label{eq:pc}
\end{align}
As the PC is a function of $p_\theta \left({\bm x}| {\bm z}^{(m)} \right)$, which is specified by the patch $m$, we simplify the notation and write $PC(m)$ instead of $PC \left( p_\theta \left({\bm x}| {\bm z}^{(m)} \right) \right)$ for the sake of brevity.

We can see that the heart of the $PC(m)$ is the token MSP $p_\theta \left(x^{k_i^*}|{\bm x}_{1\cdots i-1}, {\bm z}^{(m)} \right)$. 
Previous work has demonstrated that MSP provides valuable information about the model's confidence in its predictions~\cite{hendrycks2016baseline}. Additionally, another work utilizes MSP to detect hallucinations by identifying low-confidence outputs~\cite{chen2024inside}. As such, we adopt the $PC(m)$ and consider it to be the confidence of the VLM on its prediction of an output sequence $\bm x$. We will further elaborate on a strong correlation between extraction accuracy and the Patch Confidence in Subsect.~\ref{subsect:acc_vs_confidence}.

\subsection{PatchFinder} \label{subsect:patch_selection}
Given the patch's confidence (\ref{eq:pc}), PatchFinder's first step, Patch Size Optimization, determines an optimal patch size to divide the input image. Then, with the corresponding patches of that size, the second step Confidence-based Prediction determines the patch with the most confidence score and generates the prediction accordingly.

\begin{figure}[ht!]
    \centering
    \includegraphics[width=.99\linewidth]{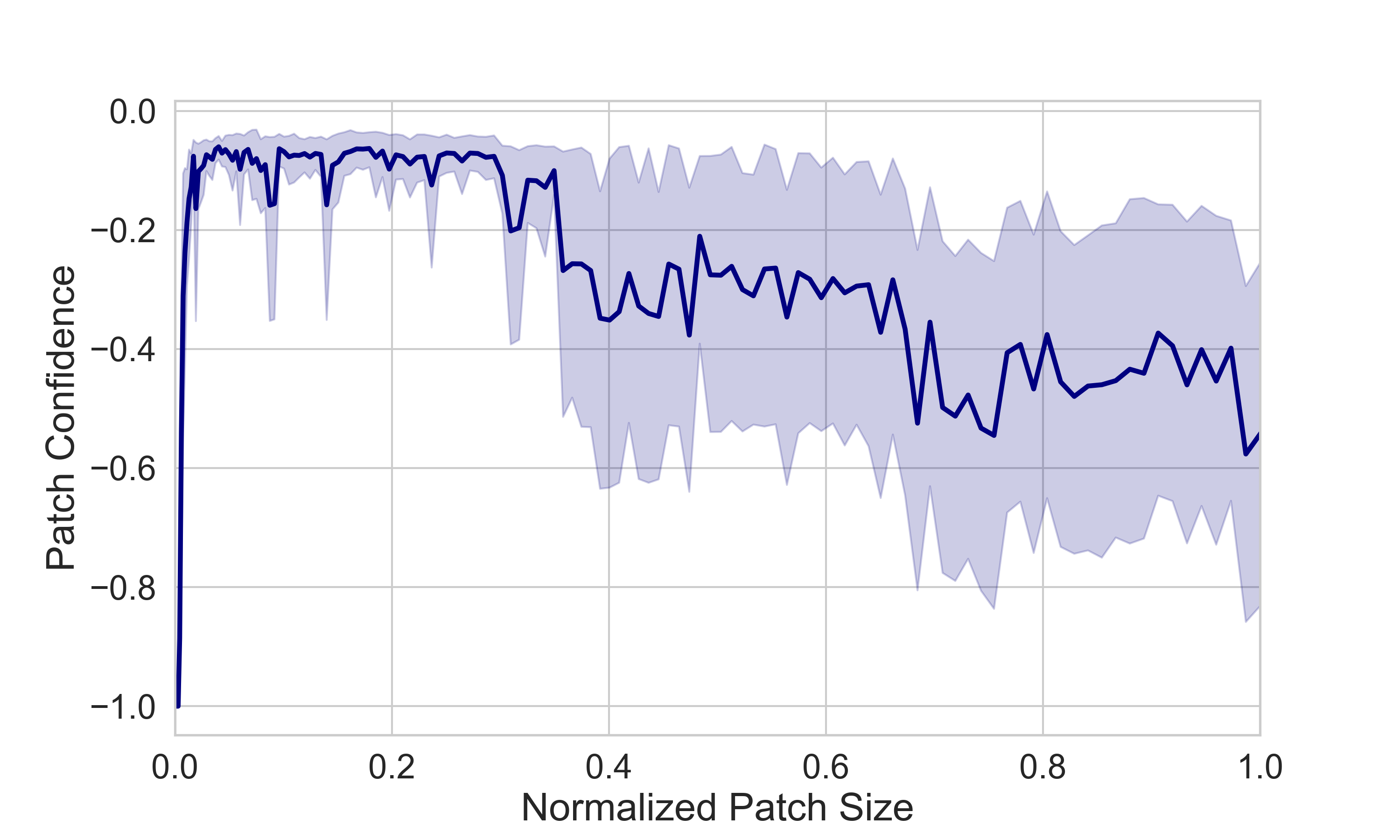}
    \caption{Patch Confidence as a function of patch size in the Colorado and Pennsylvania development split (20 documents) of the scanned well documents dataset.}
    \label{fig:patch_optimize}
\end{figure}

Patch Size Optimization determines the patch size iteratively based on the ground-truth sequences in the training data. It starts with a small initial patch size and incrementally increases it until the predicting sequence in the input images is adequately covered. Intuitively, if the patch size is too small, it will fail to cover the ground-truth sequences in the input image, resulting in low confidence and low PC scores. Conversely, if the patch size is too large, it will dilute the relevant information, also leading to lower model confidence. 
Fig.~\ref{fig:patch_optimize} demonstrates the PC as a function of the patch size in the development split of our well dataset described in \ref{subsect:exp_setting}. As we can see, using a patch that is too small results in very low confidence, while a patch that is too large leads to very high variance. Notably, there exists a significant range between $5\%$ and $30\%$ of the image that gives high-confidence predictions with low variations. In this example, we technically can select the optimal patch size to be any value between that range. It is important to note that while Patch Optimization is a beneficial step in PatchFinder, it is optional, and can be replaced by a naive patch size of 25\%.

PatchFinder then utilizes these optimally sized patches to make predictions. The process involves evaluating the PC for each patch and selecting the one with the highest confidence score. In particular, Confidence-based Prediction selects the most confident patch is determined by the maximum likelihood of that patch:
\begin{align}
    m^* = \argmax_{m \in \{1,\cdots, N(s)\} } PC(m)
\end{align}
and the estimated token $i$-th is:
\begin{align}
    \hat{x}_i = x^{k_i^*}, \ \textup{where } k_i^* = \argmax_{k \in \{1,\cdots, K\} }  p_\theta \left(x_i^k|{\bm x}_{1\cdots i-1}, {\bm z}^{(m^*)} \right)
    \nonumber
\end{align}
To sum-up, the Confidence-based Prediction step is summarized in Algorithm~\ref{algo:PatchFinder}. 

\begin{algorithm}[tb]
\caption{PatchFinder's Confidence-Based Prediction}
\label{algo:PatchFinder}
% \SetAlgoLined
\textbf{Input:} Input image $U$ and prompt query $V$ \\
 \textbf{Parameters:} Patch's size $s$  \\
 \textbf{Given:} VLM $\Phi_\theta$ \\
 \textbf{Output:} Predicted sequence $\hat{\bm x}$ 
 \begin{algorithmic}[1]
 
\STATE Divide $U$ into a set $S$ of $N(s)$ patches
\FOR{each patch $m \in S$}
\STATE $p_\theta \left({\bm x} | {\bm z}^{(m)} \right) \leftarrow \Phi_\theta(m,V)$
\STATE $PC(m) = \frac{1}{l}\sum_{i=1}^l \log \left( p_\theta \left(x^{k_i^*}|{\bm x}_{1\cdots i-1}, {\bm z}^{(m)} \right) \right) $
\ENDFOR
\STATE $m^* = \argmax_{m \in S } PC(m)$
\STATE $\hat{\bm x} \leftarrow$ maximum-likelihood from $p_\theta \left({\bm x} | {\bm z}^{(m^*)} \right)$
\RETURN $\hat{\bm x}$  

\end{algorithmic}
\end{algorithm}

\subsection{Implementation Details} \label{subsect:details}

This subsection discusses several implementation details of PatchFinder. In fact, we discuss several aspects of patch size selection, and how to improve PatchFinder with prompt engineering of the VLM as well as output filtering.

\textbf{Practical patch size selection.} The VLM model relies on positional embeddings to encode the position of tokens within the input sequence, which is crucial for capturing spatial relationships between different parts of the data. These embeddings help the model understand the relative positioning of information, allowing it to better interpret and extract target details. However, the effectiveness of these embeddings is limited by the positional encoding window within which the model was trained. When the input size is too different from this window, the model's performance degrades because it cannot effectively utilize positional information beyond its trained capacity.

Thus, the optimal patch size for a document for information extraction strongly depends on various factors, including font size, document noise level, resolution, and OCR model strength. Therefore, it is advisable to have a general understanding of the dataset and a few samples from the dataset for optimizing patch size selection as mentioned in \ref{subsect:patch_selection}. If the dataset exhibits a wide range of font sizes, a dynamic patching strategy that covers the document multiple times with different patch sizes is beneficial. In our analysis of scanned well documents, we find a patch size between $22\%$ and $25\%$ of the input document (see Fig.~\ref{fig:patch_optimize}) is beneficial for our historical well records. This selection is picked by taking the mix between the highest confidence with preference towards high patch sizes. This criteria allows the model to capture sufficient context within each patch while mitigating a significant portion of document noise.

\textbf{Prompt engineering.} When working with models like Phi-3v, which have a relatively small number of parameters compared to LLMs, we observe that they struggle to achieve high-quality inference. To address this, performing prompt engineering by referencing a subset of the dataset helps these models grasp concepts that may not have been included in their training data. For instance, in our Pennsylvania dataset, the target latitude field with the following format \textit{39\textdegree53’ 49.15”} initially posed difficulties for the model. Introducing the concept that this format represents latitude by integrating it into the beginning of the prompt was essential in improving the accuracy of these documents. In our case, we passed the following to the model: \textit{"You understand that latitude can come in a form of decimals, for example 52.25967 or in the form of degrees, for example 60\textdegree12'59.32$''$. or 60\textdegree12'59$''$."}

Furthermore, we recommend using concise and clear prompts. For example, the prompt \textit{"Extract the drilled latitude of the well described in this well completion report. Do not extract the longitude."} gives a precise instruction and clearly specifies the desired task, minimizing the risk of the model being confused by unnecessary context.

\textbf{Refinement with Output Filtering.} Our process of output filtering first involves setting a method of flagging improperly formatted responses or obvious hallucinations. Then, the mechanism removes patches corresponding to a flagged response from our pool of predictions. This refinement helps our patch confidence avoid confident hallucinations or irrelevant confident responses, and narrows down potential candidate patches. For instance, when prompting a model to produce a numerical output, we can filter out patches where the model returns a non-numerical output. 

\textbf{Model Details.} Our PatchFinder utilizes Phi-3.5-Vision model, which is a 4.2B-parameter multimodal architecture designed to process both image and text inputs to generate coherent textual outputs (Fig.~\ref{fig:patchfinder}). The image encoder leverages the CLIP ViT-L/14, a large Vision Transformer that divides images into $14\times14$ pixel patches and processes these patches as tokens using multi-head self-attentions to capture global context. The text encoder is the Llama-2's tokenizer~\cite{touvron2023llama} with vocabulary size of 32,064. The extracted visual tokens are then interleaved with text tokens and passed to the Phi-3.5-mini transformer decoder, which is a smaller variant of the larger Phi-3.5, optimized for multimodal tasks. It consists of 3.5 billion parameters, distributed across 24 transformer layers. Each layer in the decoder is equipped with multi-head self-attention mechanisms and feed-forward networks, enabling it to attend to both visual and text tokens.

\begin{figure*}[ht!]
    \centering
    \includegraphics[scale=0.27]{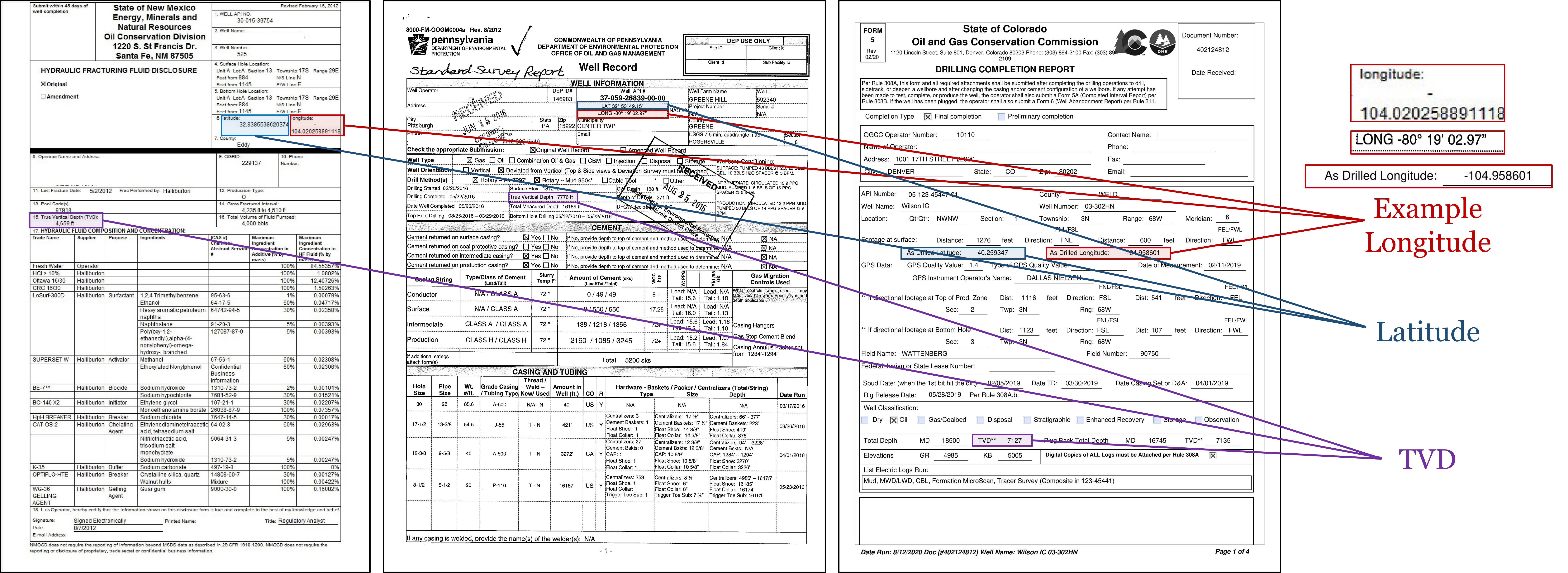}
    \caption{Examples of different drilling completion reports and well records. It can be seen that the data is not only highly different in terms of fonts, formats, and structure; their target information is also represented with different indicators.}
    \label{fig:drilling_reports}
\end{figure*}

\section{Experimental Results} \label{sect:experiments}
In this section, we report the experimental results of PatchFinder on our full historical well records, as well as test PatchFinder on noisy financial statements. We first describe the well records and elaborate on the complexity of the task in Subsect.~\ref{subsect:exp_setting}. We then provide an illustrative experiment validating the usage of PC to improve the extraction capability of VLM in Subsect.~\ref{subsect:acc_vs_confidence}. The performance of different extraction methods are finally reported in Subsect.~\ref{subsect:exp:well} and~\ref{subsect:exp:fin}.

\subsection{Dataset and Experimental Settings} \label{subsect:exp_setting}

Our dataset consists of 190 scanned historical well records from Pennsylvania (10 records), Colorado (90 records), Old Colorado (77 records) and New Mexico (13 records). These documents are part of a large national effort to find and remediate orphaned oil and gas wells \cite{omalled}. 

Each state has its own set of well record documents and these forms have changed over more than 150 years that oil and gas wells have been drilled, creating a multitude of different forms. Such a wide array of documents makes non-finetuned techniques such as PatchFinder highly valuable. Examples from these datasets are shown in Fig.~\ref{fig:drilling_reports}, which demonstrates the high variety and contrast in fonts, scan qualities, formats, and levels of noise. 

Following the study of \cite{Ma2024}, we aim to extract the TVD, latitude, and longitude from these well records. These fields allow us to locate the wells, and determine whether or not the wells intersect the water supply.

This task requires the extraction methods to have the capability to handle text of different lengths, structures and positions within the document. Given the nature of latitude and longitude, a small numerical or symbol error can lead to very significant downstream mistakes, potentially confusing one well for another. Since we are only interested in correct or incorrect responses and are performing field extraction, we use accuracy over F1 for this study.

Our team is working with state geologic surveys to expand this dataset with many more documents that have human-verified labels to enable more robust training and evaluation in the future. For our experiment, we split our 190 document dataset into 20 samples (10 from Colorado and 10 from Pennsylvania) for method development and 170 samples for testing. The 20 samples are chosen such that they represent a diverse range of document complexities and noise levels. PatchFinder does not require or use fine-tuning, thus this development split was specifically for the creation of PatchFinder as well as the Patch Optimization step.

\begin{figure}[ht!]
    \centering
    \includegraphics[width=.99\linewidth]{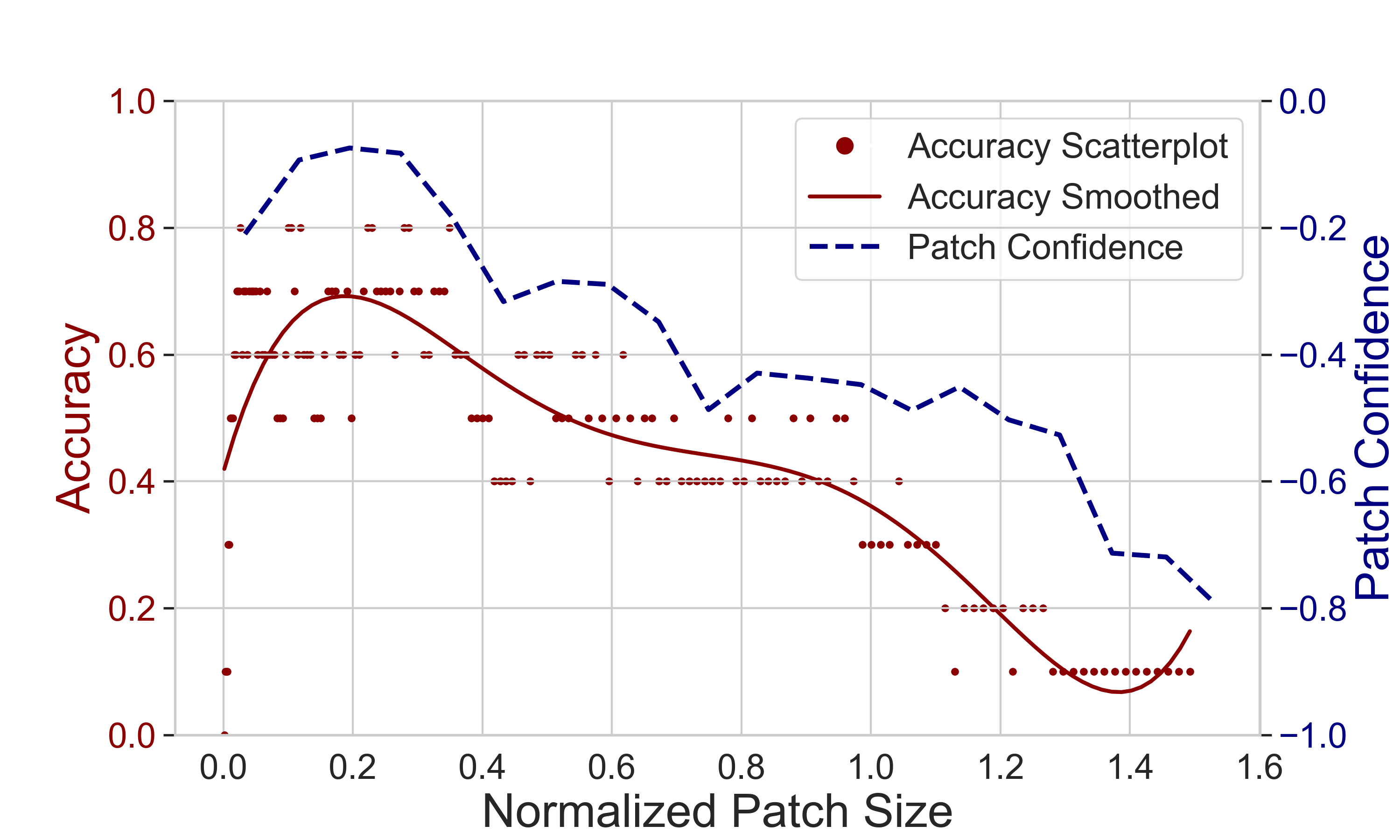}
    \caption{Correlation between Accuracy and Patch Confidence along different patch sizes.}
    \label{fig:acc_confidence}
\end{figure}

\begin{table*}[htbp]
    \centering
    \begin{tabular}{lccccc}
        \toprule
        \textbf{Model} & \textbf{Overall} & \textbf{Colorado} & \textbf{Pennsylvania} & \textbf{Old Colorado} & \textbf{New Mexico} \\ 
        \midrule
        ChatGPT-4o & 75.4\% & 100\% & 66.7\% & 77.1\% & 57.6\% \\ 
        Phi-3v & 61.8\% & 100\% & 66.7\% & 47.2\% & 33.3\% \\ 
        \textbf{Phi-3v PatchFinder} & 90.1\% & 100\% & 83.3\% & \textbf{89.2}\% & 87.9\% \\ 
        \textbf{Phi-3v Optimized PatchFinder} & \textbf{94.0}\% & 100\% & \textbf{93.3}\% & 88.7\% & \textbf{93.9}\% \\ 
        \bottomrule
    \end{tabular}
    \caption{Performance of different information extraction methods on Colorado, Pennsylvania, Old Colorado, and New Mexico documents. ChatGPT-4o tests were performed on June 20th, 2024.}
    \label{tab:exp_result}
\end{table*}

\begin{table}[htbp]
    \centering
    \begin{tabular}{lc}
        \toprule
        \textbf{Model} & \textbf{Accuracy} \\
        \midrule
       ChatGPT-4o & 36.7\% \\
        Phi-3v & 0.0\% \\
        \textbf{Phi-3v PatchFinder} & \textbf{53.3}\% \\
        \bottomrule
    \end{tabular}
    \caption{Performance of different information extraction methods on Noisy Financial Statements.}
    \label{tab:exp_result_final}
\end{table}

\subsection{Accuracy and Patch Confidence} \label{subsect:acc_vs_confidence}

As one key element of PatchFinder is in the utilization of Patch Confidence, this subsection provides an illustration on how Patch Confidence is related to the overall inference capability of the VLMs. 

To investigate this, we systematically varied the patch size, ensuring the target information remained centered, and measured the model's accuracy and confidence as functions of the patch size. By cropping the input size to fit within the positional encoding window, we ensured the model could fully leverage the positional embeddings, thereby capturing the target fields.

Our goal was to observe how the PC and the extraction accuracy change as the patch size varies from small to large values. From Fig.~\ref{fig:acc_confidence}, we can see that that for small and large  patch sizes, both the PC and the accuracy were low, indicating insufficient context and excessive noise, respectively. More importantly, both quantities achieve their optimal values around $20\%$ of the input images. This strong correlation between the PC and the extraction performance validates the usage of confidence to enhance our examined task.

\fontsize{10pt}{12pt}\selectfont
\subsection{Results on Historical \texttt well Records}
\label{subsect:exp:well}
Our initial evaluation of PatchFinder is conducted without Patch Size Optimization. In this case,  we selected an unoptimized patch size of $16.7\%$ of the total document. This preliminary approach achieved an accuracy of $90.1\%$ across our historical well records, significantly improving upon Phi-3v's accuracy of $61.8\%$.
By utilizing the Patch Size Optimization step (Subsect.~\ref{subsect:patch_selection}), which is referred to as Optimized PatchFinder, we identify the optimal patch size to be about $23\%$ of the input image.

\begin{figure*}
    \centering
    \includegraphics[height=4.2cm]{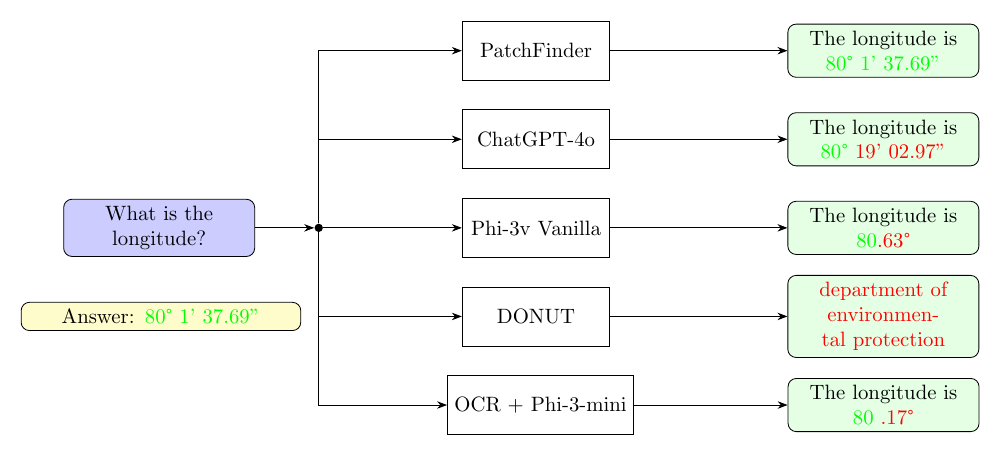}
    \caption{Qualitative Examples. Green indicates a correct token, red indicates an incorrect token.}
    \label{fig:qualex}
\end{figure*}

Using this patch size, Optimized PatchFinder achieves a significant increase in accuracy on both the Pennsylvania and New Mexico datasets, pushing accuracy up to $94\%$ across our whole dataset. Fig.~\ref{fig:qualex} depicts the qualitative difference in the responses of different models. Due to the success of our model, we used ChatGPT-4o to set up a baseline for performance. ChatGPT-4o \cite{openai2024chatgpt} is one of the premier vision language models, boasting near-top scores in many public benchmarks. When using ChatGPT for extraction tasks, it uses a multitude of techniques to increase performance, including OCR scanning, zooming, increasing image contrast, and many others.

On our dataset, ChatGPT-4o set an average accuracy of $75.4\%$ across our four document categories. A more details results for each category are shown in Table~\ref{tab:exp_result}.

\subsection{Results on Noisy Financial Statements} \label{subsect:exp:fin}
\label{NewData}
To further examine PatchFinder, we use the method to extract information in 30 historical financial statements from the University of Pennsylvania’s Historical Annual Reports Database \cite{corporate_reports_online}. These financial statements all had different layouts with similar information. For choosing the questions, we picked the most straightforward field on each document. These fields were chosen before benchmarking, as to prevent bias towards our method.  Using only one record as a reference for fixing our hyperparameters, we used rectangular patches that span the entire width of our document. This allows us to contextualize information from different sides of the document into the same patch, which is important for understanding financial reports such as these. 

For increased difficulty, we introduced a substantial amount of Gaussian noise (\(\sigma\) = 0.2). This noise was generated to be mostly white so that black text on the white background would "fade". We believe that this approach was suitable for testing the robustness of Patchfinder since documents with poor scan quality tend to have noise and/or occlusions that makes text much more difficult to read. The experimental result is shown in Table~\ref{tab:exp_result_final}, in which Patchfinder achieves a 53.3\% accuracy, significantly surpassing both ChatGPT-4o’s accuracy of 36.7\% and Phi-3v's  accuracy of 0\%. This result demonstrates that our method has potential to generalize in other settings.

\subsection{Testing PatchFinder on Known Datasets}
Common datasets and benchmarks present a way to test how generalizable PatchFinder is to diverse tasks and documents. To evaluate this, we tested PatchFinder on both the CORD and FUNSD datasets. Since PatchFinder is attached to VLM's and meant for field extraction, we constructed a simple QA dataset for CORD where the question was to extract the price of the first item on the receipt. This allowed a consistent way for us to test PatchFinder on a dataset typically meant for full extraction. Since PatchFinder does not require fine-tuning, we tested its zero-shot accuracy on all three splits of the CORD dataset, totaling 1,000 receipts. See Table \ref{tab:exp_result_fin} for results.

In testing the FUNSD dataset, we constructed a QA form by providing ChatGPT-4o with all QA fields and asking it to pick the best one for testing VLM accuracy. This resulted in a dataset of 50 different documents with both numerical and non-numerical question answer pairs. Due to the simplicity of the FUNSD dataset, both DONUT and PatchFinder were both able to get near-perfect scores. 

\begin{table}[htbp]
    \centering
    \begin{tabular}{lcc}
        \toprule
        \textbf{Model} & \textbf{CORD} & \textbf{FUNSD} \\
        \midrule
        DONUT & 46.8\% & 93.8\% \\
        Phi-3v PatchFinder & 77.2\% & 93.8\% \\
        Phi-3v PatchFinder (Upscaled) & 79.0\% & 93.8\% \\
        \bottomrule
    \end{tabular}
    \caption{Performance of PatchFinder vs. DONUT on CORD and FUNSD in terms of accuracy. (Upscaled indicates increasing the resolution of input images)}
    \label{tab:exp_result_fin}
\end{table}

\section{Limitations and Discussion} \label{sect:limit}
While PatchFinder can be used on any extraction task, our results on the CORD dataset suggest that it performs the best on highly noisy environments, and only boosts the performance of the vanilla model if there is enough noise. For the absolute best results, PatchFinder does uniquely well on extracting small strings of numbers. It is important to note that for text that spans across the whole document, it is good to change the aspect ratio of our patches. While the chunking method of PatchFinder does lower the computational costs of an individual pass, it can result in slower computation times depending on the number of patches and overlap. 

The methods used in this paper are meant specifically to rival other field extraction methods based on QA, and are not meant for extracting all information in a document. We would like to stress that the two-step OCR + LLM methods have a very similar parameter count to PatchFinder + Phi-3v. Every experiment performed for this paper was performed on a single laptop with 64G of RAM, so we believe that most individuals should be able to take advantage of PatchFinder as a way to extract information from noisy documents.

\section{Conclusion} \label{sect:conclusion}

In this study, we developed PatchFinder, a method for information extraction from noisy scanned documents. Using Phi-3v, we presented a method of Patch Optimization and confidence-based prediction, which enabled us to circumvent much of the noise present on these documents. By employing our PatchFinder method, we were able to outperform ChatGPT-4o by 18.5 percentage points on our dataset, all while being able to run our method on just a laptop. We believe that our study has shown that patching methods, and the use of MSPs as a proxy for uncertainty both have the possibility to drastically improve the performance of VLMs, and should be considered as a tool for enhancing predictions without fine-tuning. We also believe that cropping the image may be able to serve as a form of chunking, allowing models with small input lengths to consider images with higher resolutions.

\section*{Acknowledgments}
This work is supported by the Department of Energy's Undocumented Orphan Well program through the CATALOG consortium (\url{catalog.energy.gov}). LA-UR number "LA-UR-24-23837".
\vspace{1em}

\bibliographystyle{IEEEtran}

\end{document}